\documentclass[conference,a4paper]{IEEEtran}
\IEEEoverridecommandlockouts
\usepackage{cite}
\usepackage{amsmath,amssymb,amsfonts}
\usepackage{algorithmic}
\usepackage{graphicx}
\usepackage{textcomp}
\usepackage{xcolor}
\usepackage{hyperref}
\usepackage{multirow}
\usepackage{cleveref} 
\usepackage{microtype}
\usepackage{colortbl}
\usepackage{fancyhdr}

\def\BibTeX{{\rm B\kern-.05em{\sc i\kern-.025em b}\kern-.08em
    T\kern-.1667em\lower.7ex\hbox{E}\kern-.125emX}}

\usepackage{eso-pic}
\newcommand\AtPageUpperMyright[1]{\AtPageUpperLeft{
 \put(\LenToUnit{-0.08\paperwidth},\LenToUnit{-1cm}){
     \parbox{0.5\textwidth}{\raggedleft\fontsize{9}{11}\selectfont #1}}
 }}

\newcommand{\conf}[1]{
\AddToShipoutPictureBG*{
\AtPageUpperMyright{#1}
}
}
\conf{The $33^{rd}$ Asian Test Symposium (ATS 2024)} 

\begin{document}
\title{
Evaluating Different Fault Injection Abstractions on the Assessment of DNN SW Hardening Strategies
\vspace{-2.5mm}
}

\author{ 
\IEEEauthorblockN{ 
    Giuseppe Esposito, Juan-David Guerrero-Balaguera,
    Josie E. Rodriguez Condia,
    Matteo Sonza Reorda
  } 
  
    \IEEEauthorblockA{Politecnico di Torino - Department of Control and Computer Engineering (DAUIN), Turin, Italy\\
    \{giuseppe.esposito, juan.guerrero, josie.rodriguez, matteo.sonzareorda\}@polito.it
    }
    \vspace{-10mm}
}
\vspace{-3.5mm}
\newcommand\todo[1]{\textcolor{red}{#1}}

\IEEEoverridecommandlockouts \IEEEpubid{\makebox[\columnwidth]{979-8-3315-2916-1/24/\$31.00 \copyright2024 IEEE \hfill} \hspace{\columnsep}\makebox[\columnwidth]{ }}

\maketitle

\IEEEpubidadjcol

\begin{abstract}\footnote{This work has been supported by the National Resilience and Recovery Plan (PNRR) through the National Center for HPC, Big Data and Quantum Computing.\vspace{-7.5mm}} The reliability of Neural Networks has gained significant attention, prompting efforts to develop SW-based hardening techniques for safety-critical scenarios. However, evaluating hardening techniques using application-level fault injection (FI) strategies, which are commonly hardware-agnostic, may yield misleading results. This study for the first time compares two FI approaches (at the application level (APP) and instruction level (ISA)) to evaluate deep neural network SW hardening strategies. 
Results show that injecting permanent faults at ISA (a more detailed abstraction level than APP) changes completely the ranking of SW hardening techniques, in terms of both reliability and accuracy. These results highlight the relevance of using an adequate analysis abstraction for evaluating such techniques.
\end{abstract}

\begin{IEEEkeywords}
Deep Neural Networks, Fault tolerance, SW hardening, Reliability assessment, Safety, Artificial Intelligence
\end{IEEEkeywords}
\section{Introduction}
    Today, Artificial Intelligence (AI) influences many areas of technology. It is applied in IoT devices, like robots, drones, and automotive systems, as well as chatbots across various domains powered by Large Language Models. The complex nature of AI requires significant computational resources, usually provided by powerful hardware systems supported by specialized AI accelerators, including \textit{Graphics Processing Units} (GPUs)~\cite{libri2020paella, nvidia-Drive, AMD_Instinct_2024}. In fact,
    the development of AI accelerators (e.g., GPUs) is boosted by the continuous evolution of semiconductor technologies. 
    Unfortunately, as the semiconductor technologies scale down (7nm and below), devices exhibit increased susceptibility to faults, compromising system reliability. Some studies indicate that factors, such as process variation, wear out, and harsh environments accelerate device aging and degradation, resulting in hardware faults that manifest as \textit{Silent Data Errors} (SDEs) \cite{dixit2021silent, hochschild2021cores, singh2023silent} at the application level. In AI applications (e.g., \textit{Deep Neural Networks}, or DNNs), faults affecting the hardware can lead to computational errors/failures, reducing their performance and compromising their correct behavior. 
    
    Several works aimed at reducing the impact of hardware faults on DNNs by adopting software-based fault mitigation strategies acting on the DNN architecture~\cite{9897813}. These fault countermeasures seek to increase the robustness or, in other cases, complement well-known approaches, such as error-detection/-correction mechanisms (i.e., ECCs) \cite{correia2012practical, peterson1972error}. In this work, we target only software-based hardening techniques implemented at the application level, neglecting lower-level software-based hardening strategies (e.g., Algorithmic-Based Fault Tolerance techniques \cite{stefanidis2004algorithm}). Among the software-based hardening approaches for DNNs, the most common ones resort to activation bounding with or without retraining (e.g., \textit{Ranger}\cite{chen2021low}, \textit{Adaptive Clipper}\cite{SC-hardening}, \textit{Swap ReLU6}\cite{9897813} or \textit{Median filter}\cite{ozen2020boosting}), redundancy (e.g., TMR) or hardware-aware pruning for TPU architectures \cite{abdullah2020salvagednn}. The experiments supporting the effectiveness of these methods generally resort to Fault Injection (FI) campaigns that inject permanent or transient faults to validate the different techniques. 

    These FI campaigns can be carried out at different levels of abstraction, such as \textit{i)} \textit{physical-based}, \textit{ii)} \textit{hardware-based}, \textit{iii)} \textit{Instruction/ISA-based}, and \textit{iv)} \textit{Application/APP-based}~\cite{gnad_2024}. The first three categories are hardware aware and can provide realistic results; nonetheless, they require specialized tools and hardware descriptions, leading to costly, nonscalable, and prohibitive evaluation times, especially when dealing with DNN workloads~\cite{ruospo2021pros}. In contrast, application/APP-based FI strategies are more flexible and faster than other approaches. Consequently, the reliability evaluation of DNNs (and their SW-based hardening strategies) has widely adopted FI evaluations at the APP level (i.e., corrupting DNN weights/feature maps).
        Unfortunately, evaluations at the APP level are hardware-agnostic, so they can hardly describe the actual effect of faults occurring in the underlying hardware, and this affects the accuracy of the evaluations. As an alternative solution, hardware injection through program transformation (HITPT) is a SW-based FI technique that corrupts at the instruction level (ISA) the execution of any parallel program, including DNNs. 
    To do so, HITPT modifies the assembly code of the DNN execution to inject a fault in the target component. 
    HITPT allows modeling the effect of permanent faults in a real device more realistically, which is suitable for effective reliability evaluation of complex applications, such as DNN workloads and SW hardening strategies. 
    Moreover, HITPT is faster than evaluation-based FI approaches and provides equivalent insights into faults affecting the GPUs as the FI occurs closer to the internal HW structures than application-level approaches. 

    In this work, we present, for the first time, a comparative evaluation made by resorting to different FI abstractions to study the effectiveness of several DNN SW hardening strategies w.r.t. permanent faults. The comparison is made by resorting to custom tools:  a customized version of PytorchFI \cite{Mahmoud_2020}, which allows the corruption of weights and feature maps of a DNN, and 
    an adapted version of NVBitFI, which injects permanent faults (i.e., stuck-at faults) in the register files and the input/outputs of the functional units of a GPU. We considered 3 general-purpose DNNs (\textit{Lenet5}, \textit{MobilenetV2}, and \textit{Resnet18}) in the original and hardened versions, resorting to 3 different state-of-the-art SW hardening strategies (\textit{Swap ReLU6} \cite{9897813}, \textit{Ranger} \cite{chen2021low} and \textit{Adaptive Clipper} \cite{SC-hardening}). The results show that results produced by APP-level FI are significantly different than those from ISA-level FI. More in general, the adoption of a hardware-aware FI technique (thereby providing more accurate results) may result in a complete change of the ranking of the SW hardening techniques w.r.t. the ranking obtained resorting to APP-level FI. 


    The paper is structured as follows: Section II outlines the possible reliability evaluation strategies. Section III describes the experimental environment and the supported Fault Models. The experimental setup and the results are reported in Section IV. Section V concludes the paper and lists possible future works. 

    \section{Related works}
        \label{sec:rel_work}
    \begin{figure}
            \centering
            \includegraphics[width=\linewidth]{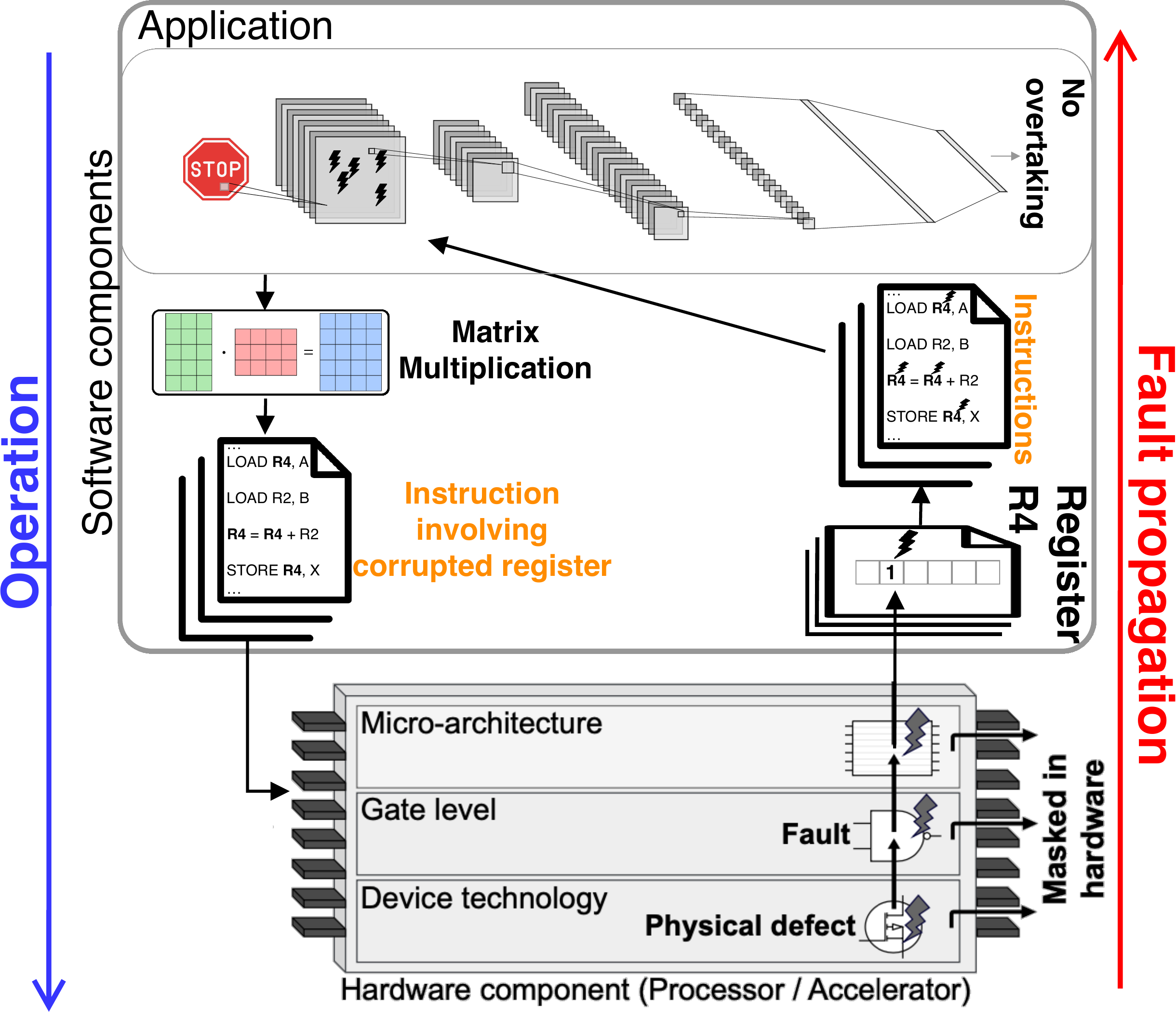}
            \caption{Permanent fault propagation from the hardware to the application possibly generating an error. 
            }
            \label{fig:fault_propagation}
            \vspace{-7mm}
        \end{figure}

    Over recent years, several studies have addressed the challenge of effectively assessing DNN robustness through FI techniques, focusing on evaluation strategies that can work at different levels of abstraction such as \textit{i)} APP, \textit{ii)} ISA, \textit{iii)} microarchitectural-level and \textit{iv)} Register Transfer Level (RTL). The \textit{APP} FI campaigns involve the perturbation of the parameters of the DNN or of the intermediate outputs. For example, authors of \cite{9217880, malekzadeh2021impact} conducted 2 different reliability-oriented analyses evaluating the best DNN weights representation format and the best DNN architecture, respectively, by performing random bit-flips on weight parameters. The approach simulates errors caused by faults without accounting for how data and operations are assigned to the hardware. Although the high customization degree and low execution time represent advantages in terms of controllability and feasibility of the experiments, a key drawback of APP FIs is the hardware-agnostic approach that is adopted \cite{ruospo2021pros}. 

    \textit{ISA} FIs provide a better trade-off between evaluation time and result accuracy. ISA FI can resort to the HITPT strategy, which allows for injecting faults in kernel instructions of parallel executions \cite{tsai2021nvbitfi}. For transient faults on GPUs, NVBitFI \cite{tsai2021nvbitfi} and SASSIFI \cite{hari2017sassifi} are state-of-the-art frameworks based on the corruption of the binary code and the PTX code, respectively, to inject faults. DNNs reliability assessment through ISA-level FIs has been explored in \cite{9946344, guerrero2022reliability} where the authors extended the use of this approach to permanent faults affecting registers and functional units. 
    Moreover, in \cite{condia2022multi} the authors identified the input values that activate faults injected by instrumenting the GPU assembly code. Lastly, in \cite{papadimitriou2021demystifying} the authors conducted an analysis of two ARM ISAs and two microarchitectures for each ISA to quantify errors in architecture-level and SW-level vulnerability evaluation. 
    
    
    Fine-grained reliability evaluations (at the \textit{micro-architectural level}) of DNNs on hardware like GPUs are highly detailed but extremely time-consuming. 
    For example, GUFI \cite{tselonis2016gufi} performs FIs on a GPU microarchitecture simulator, corrupting hardware components at random execution cycles rather than altering the application source/binary code. Despite higher experimental accuracy, GUFI's long evaluation times make it impractical for complex applications like DNNs.

    \textit{RTL} abstraction, which describes circuit behavior through data transfers between registers and logical gates, allows for more detailed hardware characterization. For example, RTL FIs were used in \cite{condia2022effective} to identify critical GPU sites and implement selective hardening for improved reliability. However, the high resource consumption, complexity, and huge development time make a full reliability assessment of DNNs impractical at this level (or lower).

    To summarize, most of the current DNN reliability evaluation approaches adopt the APP FI due to its easy applicability and high speed, but neglect the result inaccuracy. To the best of our knowledge, no existing study has analyzed the effect of different abstraction levels when evaluating SW hardening strategies for DNNs. This work for the first time compares the results coming from ISA FIs and APP FIs when SW hardening strategies are applied to DNNs.  

\section{Evaluation methodology}
    
    This section describes the adopted strategy to evaluate and analyze the impacts of two different abstraction levels (ISA and APP) on the reliability assessment of SW-hardening techniques for DNNs. In particular, we evaluate permanent (stuck-at) faults during the inference of DNNs on GPUs.
    The first abstraction (ISA or \textit{CUDA instruction-level}) focuses on the HITPT mechanisms to corrupt the program, while APP evaluation induces errors by directly corrupting DNN parameters to represent errors from hardware faults. 
    Based on the error induced on the final DNN output, the fault classification allowed us to outline fault effects for each hardened application, highlighting its resiliency in front of different abstraction levels. Moreover, the performed FI experiments allow assessing the severity of the injected faults and the consequent induced error. 

    \subsection{Hardware-Injection Through Program Transformation (HITPT)}
        \label{sec:CUDA}
        As illustrated in \ref{fig:fault_propagation}, when an operation is executed, defects might be activated and propagated as a hardware fault (e.g., corrupting the logic state on a gate) across the micro-architecture of the system during the application's execution and cause failures/errors in the system. More in detail, corrupted logic states can impact multiple operations of assembly instructions (e.g., corrupted assembly code workflow) and produce incorrect final application outputs or impact their parameters.
        
        To inject faults, in this work, we employ an adapted version of NVBitFI, which implements the HITPT approach to model, at the SW level, the fine-grain effect of faults in the data path structures of a GPU. In particular, the framework injects permanent (\textit{stuck-at}) faults in the inputs/outputs of Functional Units (FUs) and on Register File (RFs) destinations for the kernel instructions. Then, we perform exhaustive FI campaigns on all used registers per RF and FUs (separately) for a targeted Thread, which is executed on a specific \textit{Streaming Multiprocessor} (SM) inside a GPU. 

        
       
    \subsection{Application-level FI}
        \label{sec:AL_FI}
        The fault injector used for this type of abstraction level is based on PyTorchFI \cite{Mahmoud_2020}, which allows the corruption of DNN components on the parameter tensor (\textit{weights}) or in the intermediate output tensor (\textit{neuron's output}).

        For each DNN under testing, we have performed 2 FI campaigns. In the first one, we injected Single Bit-Flips (SBFs) for each DNN run through statistical FI \cite{10136998} targeting weight parameters. In the second campaign, we injected Multiple Bit-Flips (MBFs) according to a given BER.
        In the first part of the campaign, we targeted weight parameters and the injected faults are obtained by extracting a BER fraction out of the available bit space. After sampling a sufficient number of faults, using the PytorchFI forward hook function, we can perturb the weights before the DNN inference. 
        While the DNN weights are fixed, the neuron's output changes depending on the target layer's input \textit{Feature Map} (FM). Therefore, in the second campaign, we targeted neurons' outputs through PytorchFI which allows to corrupt DNN intermediate output during inference. We used the framework presented in \cite{10538938} where the error list contains the desired tail size the Block Error Rate ($BlER$), the Neuron Error Rate ($NER$), and the bit location. All those fault features, define the FM portion to corrupt along with the specific bit locations. 
        
    
    \subsection{Evaluation process}
        
        For each level of FI abstraction, injected faults are classified into different categories depending on the severity of their impact on the application performance. Specifically, the recorded accuracies of the fault-free and corrupted DNNs are evaluated so that the framework can compute the \textit{Relative Accuracy Degradation} (RAD) to quantify the DNN accuracy degradation. Mathematically, RAD is defined as $\frac{ACC_{fault-free} - ACC_{faulty}}{ACC_{fault-free}}$ where $ACC_{fault-free}$ represents the golden accuracy and $ACC_{faulty}$ is the accuracy of the corrupted DNN. The corresponding prediction confidences are compared with the ones generated in the fault-free scenario and the fault classifier assigns the label \textit{Masked} if they match (because the injected error does not change the inference output), or \textit{Safe Silent Data Corruption (Safe-SDC)} (when the fault does not change the prediction),  or \textit{Critical Silent Data Corruption (Critical-SDC)} (when the fault does change the prediction). In case the fault has produced a system hang or crash, it is classified as \textit{Detectable Unrecoverable Error (DUE)} that is generally due to memory access violation or memory misalignment violation. 

        In this work, we first perform a preliminary assessment comparing the fault distribution of the campaigns conducted at ISA (injecting single stuck-at faults) and APP (performing SBFs) abstraction levels to compare the effect of the fault/error injected in 2 different levels of abstraction.
        
        Nevertheless, from a SW standpoint, the propagation of the error, induced by corrupting assembly operations, up to the DNN execution can turn into the corruption of several intermediate outputs/parameters. Therefore, to make a fair comparison, we have computed the error induced by single stuck-at faults at ISA and we induced the same error at APP to make a fair comparison between 2 abstractions. To do so, after each ISA FI, we count and preserve the number of times that the fault is excited (i.e., the target bit assumes a different polarity, either 0 or 1, w.r.t. the type of injected stuck-at fault) by the CUDA kernel execution. Consequently, by dividing by the total number of times the register is used, we have computed the induced Bit Error Rate (BER). After performing ISA FI campaigns, the results obtained were used to define the range [$BER_{min}$,$BER_{max}$]. Subsequently, 10 values were sampled within this range and employed to perform MBFs at APP. After running the simulations, we compared the impact (in terms of accuracy) of FI campaigns at different abstraction levels.
        This enabled a fair comparison in the accuracy of the benchmarks under identical error activation rates.

\section{Experimental results}
    \label{sec:exp_res}
    
    The first set of FI campaigns resorts to ISA FI and focuses on permanent (\textit{stuck-at}) faults arising on all registers in the RFs and all FU cores from the first SM and targeting a given Thread ID. In detail, we injected all possible single faults in registers (totaling 10,496 runs) and all possible single faults in the inputs and outputs of the FUs (1,536 runs). 

    For the second FI campaign (at the APP level), we calculated the optimal number of corruptions (i.e., single stuck-at faults are injected in the weights as presented in~\cite{9217880}) such that experiment confidence level =95\% and error margin=0.5. When multiple bit-flips for each evaluation are injected, the percentage of faults (i.e., BER) ranges within $[1\cdot 10^{-6}, 6\cdot 10^{-4}]$. 

    \begin{figure*}
            \centering
            \includegraphics[width=0.94\textwidth]{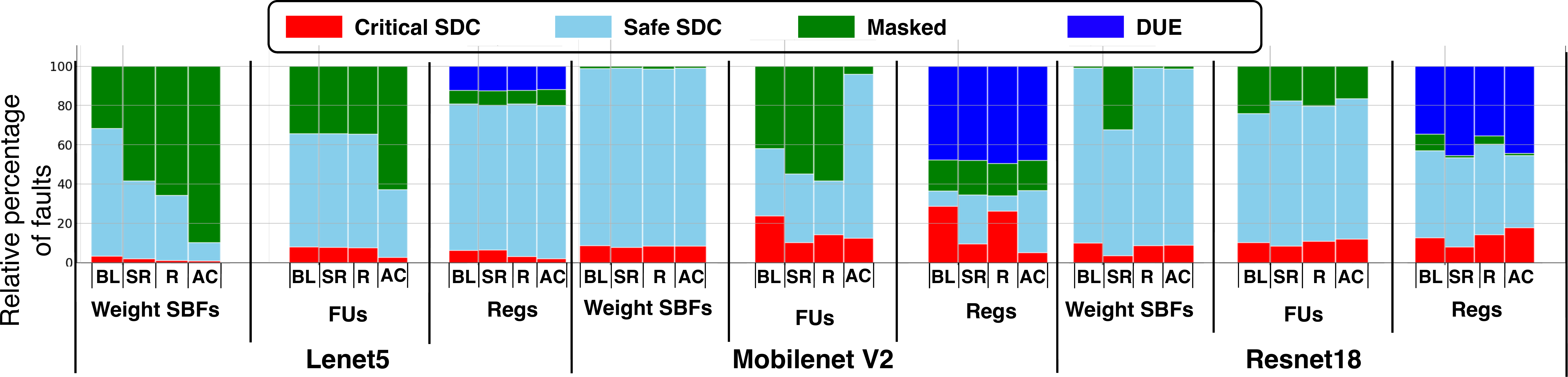}
            \vspace{-4 mm}
            \caption{Fault distribution for 3 DNNs where FI campaigns target \textit{i)} weight parameters by performing single bit-flips (Weight SBFs) at the APP, \textit{ii)} Functional Units (FUs) at ISA and \textit{iii)} Registers (Regs) at ISA. Specifically, the proportion of faults is indicated in relation to each compromised component. Moreover, each bar corresponds to the baseline  (BL) unhardened model, Swap ReLU6 (SR), Ranger (R) and Adaptive Clipper (AC).
            }
            \label{fig:LeNet_FI}
            \vspace{-6 mm}
        \end{figure*}

    The ISA FI experiments were performed on a workstation HP Z2 G5 with an Intel Core i9-10800 CPU with 20 cores, 32 GB of RAM, and equipped with one NVIDIA GPU RTX 3060TI, requiring around 96h ($\approx60h$ for Regs FIs and $\approx36h$ for FUs FIs). On the other hand, the APP FIs used a 6-node cluster system with 2 Intel 16-core Xeon Scalable Processors Gold 6130 2.10 GHz per node and equipped with 6 NVIDIA V100 SXM2 and 32 GB of RAM, requiring around 3h ($\approx 1h$ for SBFs FIs and $\approx2h$ for MBFs FIs). For both FI campaigns (ISA and APP), we considered three SW-based hardening techniques (\textit{Ranger} \cite{chen2021low}, \textit{Swap ReLU6} \cite{9897813}, and \textit{Adaptive Clipper} \cite{SC-hardening}) applied on \textit{LeNet5} (trained on Mnist), \textit{Mobilenet V2}, and \textit{Resnet18} (both trained on Cifar10) and image classifiers, as representative DNN workloads.

    \subsection{Error distribution impacts from both evaluation abstractions}
    \label{sec:only_single}

    Fig. \ref{fig:LeNet_FI} outlines the relative fault distribution stemming from the APP (\textit{weight SBFs}) and ISA (\textit{FUs} and \textit{Regs}) evaluations on the three analyzed hardened-free DNNs (\textit{baseline} or \textbf{BL}) and their hardened versions (\textit{Swap ReLU6} or \textbf{SR}, \textit{Ranger} or \textbf{R}, and \textit{Adaptive Clipper} or \textbf{AC}).

    The experimental results show that, according to the reliability assessment abstraction (APP or ISA), the DNN resiliency can quantitatively differ. 
    In general, we observed that depending on the FI abstraction level the effectiveness of a hardening mechanism may change significantly. In two of the three analyzed DNNs (Lenet5 and Mobilenet V2), all FI campaigns show a clear reduction in Critical SDCs (from 28.60\% to 4.85\%). On the other side, moving to Resnet18, we notice that APP FI results show a negligible impact by the hardening techniques, while ISA FI results show that both R and AC hardening techniques produce an increase in the percentage of Critical SDCs. We can also notice that APP FI never produces any DUE, while ISA FI does when we inject into Regs. In this case, DUEs are mainly due to the corruption of instruction destination registers, leading to illegal memory accesses.     
    We can also see that ISA FI strategies significantly change (up to $3 \times$) the percentage of Critical SDCs produced by the baseline DNN inference (BL) w.r.t. the results of APP FI when Mobilenet V2 is evaluated.
    On the other side, for the baseline Lenet5 the Critical SDC percentage increases by almost $2\times$ depending on the FI approach. For Resnet18 the same percentage remains constant due to the inherent and higher redundancy of this DNN model.


    
    

    Our results indicate that the percentage of Critical SDCs produced by each DNN may change significantly (both in the BL and in any of the hardened versions) depending on the usage of Regs and FUs done by each  DNN. On the one hand, DNNs actively using registers (\textit{Mobilenet V2} and \textit{Resnet18}) show a higher percentage of Critical SDCs w.r.t. faults impacting functional units (around 2\%). On the other hand, the results on \textit{Lenet5}, with reduced utilization of accessible registers, showed a more resilient behavior to corruptions on the registers (4.41\% of critical SDCs) than on functional units (6.42\% SDCs).


    These results demonstrate that high-level FI approaches  (i.e., APP FIs) tend to overestimate the effectiveness of SW hardening techniques in terms of Critical SDCs. In fact, APP FIs report a limited percentage of Critical SDCs (no more than 10\%). In contrast, ISA FIs report a much higher percentage of Critical SDCs (up to 31\% in the execution of more complex DNNs, such as \textit{Mobilenet V2} and \textit{Resnet18}). Moreover, APP FIs completely neglect the occurrence of DUEs, which may amount to up to 47.5\% in some cases (faults in Regs in MobilNet). 

    Furthermore, it can be noticed that when weight parameters are corrupted with single stuck-at faults, all SW hardening strategies provide beneficial effects in the Lenet5 architecture, increasing the percentage of Masked faults from about 30\% up to about 90\%. Specifically, Adaptive Clipper outperforms the other techniques when it is implemented on Lenet5. 
    
        Interestingly, Adaptive Clipper reaches better performance when FUs are corrupted, decreasing the Critical faults by 5.27\% w.r.t. the baseline DNN. 
        
    
    Adaptive Clipper increases the percentage of Masked faults in Lenet5 by fine-tuning weights and bounding activation functions but is less effective when Registers are corrupted only improving the baseline Masked faults by 1.48\%. On the other hand, Mobilenet V2, with built-in range restrictions, shows lower benefits from Adaptive Clipper due to its inherent output variability and architecture in weights and FU corruptions.
    This can be evinced from the increased Safe SDCs (of 39.10\%) and decreased Masked faults (of 37.70\%) in front of a decrease of Critical SDCs (of 29.43\%) during FUs FI evaluations. From the results obtained from Regs evaluations, the decrease by 23.5\% of Critical SDCs suggests improved performance. Nevertheless, there is a slight increase in DUEs (0.10\%) compared to the baseline, indicating the types of errors the hardening technique can mask. Ranger and Swap ReLU6 show better performance by slightly reducing faults with critical effects to 14.25\% and 10.22\%, respectively, when FUs are corrupted. For Resnet18, APP evaluations show that Swap ReLU6 improves performance by reducing the Critical SDCs by 2\%, while Ranger maintains it around 10\%. Adaptive Clipper, however, is less effective, increasing Critical SDCs by 5.2\%. Resnet18 also experiences a significant rise in DUEs with Swap ReLU6 and Adaptive Clipper, by 10.09\% and 11.07\%, respectively. Swap ReLU6 reduces Critical SDCs to 7.95\%, indicating a lower error probability compared to the baseline and other strategies. In contrast, Adaptive Clipper does not improve but indeed degrades the original reliability, with Critical faults either remaining constant or increasing. This behavior is reflected in all FI results, where the reported percentage of Critical faults is kept constant (to 8.87\% in weights parameters corruption) or increased (of 2.16\% in FUs corruption). Our experiments indicate that the evaluation abstraction quantitatively affects the effectiveness of a hardening strategy (from 3\% to 0.4\% of the faults evaluated on the selected workloads). The distribution of critical faults across hardening strategies shows the same trend only for Regs and FUs, revealing different Critical SDC coverage rankings across abstraction levels. Under Regs and FUs corruptions, we can notice that Adaptive Clipper outperforms the other hardening strategies, decreasing the baseline Critical fault percentage up to 29.43\% when it is implemented on Mobilenet V2. On the other hand, APP FI highlights that Swap ReLU6 is the best-performing technique in terms of Critical fault decrease of 0.4\%. Moreover, the presence of DUEs is shown only by injecting in Regs.


    As a conclusion, the experimental results suggest that APP FIs provide misleading results when assessing DNNs and the corresponding SW hardening techniques. Instead, ISA FIs, which support a more detailed representation of hardware faults and their effects, do provide quite different results, significantly changing the hardening techniques ranking. Since ISA FIs are closer to the underlying hardware architecture, these results are indeed closer to the real ones.

    

    
\subsection{Impact of faults on DNN accuracy}
%
\begin{table*}[]
\centering
\caption{Accuracy for the baseline DNNs and their hardened versions. The simulation results are obtained by corrupting DNNs (BER, APP-level) and basic instructions (Single stuck-at, ISA-level). A color scale is used to rank the results provided by each FI campaign on the different applications. Red, orange, yellow and green color indicate the highest to the lowest accurate result for each scenario, respectively.}
\vspace{-3mm}
\begin{tabular}{c|ccc|ccccc|cccc}
\hline
\textbf{} &
  \multicolumn{3}{c|}{\textbf{Fault model}} &
  \multicolumn{5}{c|}{\textbf{BER (APP-level)}} &
  \multicolumn{4}{c}{\textbf{Single stuck-at (ISA-level)}} \\ \hline
\multicolumn{1}{l|}{} &
  \multicolumn{3}{c|}{\textbf{Hardening Strategy}} &
  \multicolumn{1}{c|}{Baseline} &
  \multicolumn{1}{c|}{\begin{tabular}[c]{@{}c@{}}Adaptive \\ Clipper\end{tabular}} &
  \multicolumn{1}{c|}{\begin{tabular}[c]{@{}c@{}}Swap \\ ReLU6\end{tabular}} &
  \multicolumn{1}{c|}{Ranger} &
   &
  \multicolumn{1}{c|}{Baseline} &
  \multicolumn{1}{c|}{\begin{tabular}[c]{@{}c@{}}Adaptive \\ Clipper\end{tabular}} &
  \multicolumn{1}{c|}{\begin{tabular}[c]{@{}c@{}}Swap \\ ReLU6\end{tabular}} &
  Ranger \\ \cline{1-8} \cline{10-13} 
\multicolumn{1}{l|}{} &
  \multicolumn{3}{c|}{\textbf{Golden accuracy}} &
  \multicolumn{1}{c|}{81.28\%} &
  \multicolumn{1}{c|}{76.40\%} &
  \multicolumn{1}{c|}{82.46\%} &
  \multicolumn{1}{c|}{81.28\%} &
  \multirow{-2}{*}{} &
  \multicolumn{1}{c|}{81.28\%} &
  \multicolumn{1}{c|}{84.13\%} &
  \multicolumn{1}{c|}{72.80\%} &
  81.28\% \\ \cline{2-13} 
\multicolumn{1}{l|}{} &
  \multicolumn{1}{c|}{} &
  \multicolumn{1}{c|}{} &
  $10^{-6}$ &
  \multicolumn{1}{c|}{\cellcolor[HTML]{9AFF99}71.24\%} &
  \multicolumn{1}{c|}{\cellcolor[HTML]{FFCCC9}55.63\%} &
  \multicolumn{1}{c|}{\cellcolor[HTML]{FFCE93}66.20\%} &
  \multicolumn{1}{c|}{\cellcolor[HTML]{FFFFC7}71.60\%} &
   &
  \multicolumn{1}{c|}{\cellcolor[HTML]{FFFFC7}72.99\%} &
  \multicolumn{1}{c|}{\cellcolor[HTML]{FFCE93}72.18\%} &
  \multicolumn{1}{c|}{\cellcolor[HTML]{FFCCC9}71.59\%} &
  \cellcolor[HTML]{9AFF99}77.96\% \\ \cline{4-8} \cline{10-13} 
\multicolumn{1}{l|}{} &
  \multicolumn{1}{c|}{\multirow{-2}{*}{\textbf{Neurons}}} &
  \multicolumn{1}{c|}{} &
  $10^{-5}$ &
  \multicolumn{1}{c|}{\cellcolor[HTML]{FFFFC7}67.36\%} &
  \multicolumn{1}{c|}{\cellcolor[HTML]{FFCCC9}51.28\%} &
  \multicolumn{1}{c|}{\cellcolor[HTML]{FFCE93}58.66\%} &
  \multicolumn{1}{c|}{\cellcolor[HTML]{9AFF99}68.08\%} &
  \multirow{-2}{*}{\textbf{Regs}} &
  \multicolumn{1}{c|}{-} &
  \multicolumn{1}{c|}{-} &
  \multicolumn{1}{c|}{-} &
  - \\ \cline{2-2} \cline{4-13} 
\multicolumn{1}{l|}{} &
  \multicolumn{1}{c|}{} &
  \multicolumn{1}{c|}{} &
  $10^{-6}$ &
  \multicolumn{1}{c|}{\cellcolor[HTML]{FFCE93}76.44\%} &
  \multicolumn{1}{c|}{\cellcolor[HTML]{FFCCC9}73.75\%} &
  \multicolumn{1}{c|}{\cellcolor[HTML]{9AFF99}84.12\%} &
  \multicolumn{1}{c|}{\cellcolor[HTML]{FFFFC7}76.88\%} &
   &
  \multicolumn{1}{c|}{\cellcolor[HTML]{FFCCC9}58.34\%} &
  \multicolumn{1}{c|}{\cellcolor[HTML]{FFFFC7}66.24\%} &
  \multicolumn{1}{c|}{\cellcolor[HTML]{FFCE93}62.62\%} &
  \cellcolor[HTML]{9AFF99}75.15\% \\ \cline{4-8} \cline{10-13} 
\multicolumn{1}{l|}{\multirow{-5}{*}{\textbf{Mobilenet V2}}} &
  \multicolumn{1}{c|}{\multirow{-2}{*}{\textbf{Weights}}} &
  \multicolumn{1}{c|}{\multirow{-4}{*}{\textbf{BER}}} &
  $10^{-5}$ &
  \multicolumn{1}{c|}{\cellcolor[HTML]{FFCCC9}66.56\%} &
  \multicolumn{1}{c|}{\cellcolor[HTML]{FFFFC7}72.45\%} &
  \multicolumn{1}{c|}{\cellcolor[HTML]{9AFF99}81.63\%} &
  \multicolumn{1}{c|}{\cellcolor[HTML]{FFCE93}69.37\%} &
  \multirow{-2}{*}{\textbf{FUs}} &
  \multicolumn{1}{c|}{\cellcolor[HTML]{FFCCC9}34.32\%} &
  \multicolumn{1}{c|}{\cellcolor[HTML]{FFCE93}49.11\%} &
  \multicolumn{1}{c|}{\cellcolor[HTML]{FFFFC7}60.63\%} &
  \cellcolor[HTML]{9AFF99}68.47\% \\ \hline
 &
  \multicolumn{3}{c|}{\textbf{Golden accuracy}} &
  \multicolumn{1}{c|}{88.14\%} &
  \multicolumn{1}{c|}{84.13\%} &
  \multicolumn{1}{c|}{72.80\%} &
  \multicolumn{1}{c|}{88.14\%} &
   &
  \multicolumn{1}{c|}{88.14\%} &
  \multicolumn{1}{c|}{84.13\%} &
  \multicolumn{1}{c|}{72.80\%} &
  88.14\% \\ \cline{2-13} 
 &
  \multicolumn{1}{c|}{} &
  \multicolumn{1}{c|}{} &
  $10^{-6}$ &
  \multicolumn{1}{c|}{\cellcolor[HTML]{FFFFC7}76.81\%} &
  \multicolumn{1}{c|}{\cellcolor[HTML]{FFCE93}76.14\%} &
  \multicolumn{1}{c|}{\cellcolor[HTML]{FFCCC9}71.21\%} &
  \multicolumn{1}{c|}{\cellcolor[HTML]{9AFF99}80.57\%} &
   &
  \multicolumn{1}{c|}{\cellcolor[HTML]{9AFF99}87.08\%} &
  \multicolumn{1}{c|}{\cellcolor[HTML]{FFCE93}81.31\%} &
  \multicolumn{1}{c|}{\cellcolor[HTML]{FFCCC9}71.75\%} &
  \cellcolor[HTML]{FFFFC7}81.62\% \\ \cline{4-8} \cline{10-13} 
 &
  \multicolumn{1}{c|}{\multirow{-2}{*}{\textbf{Neurons}}} &
  \multicolumn{1}{c|}{} &
  $10^{-5}$ &
  \multicolumn{1}{c|}{\cellcolor[HTML]{FFCE93}73.99\%} &
  \multicolumn{1}{c|}{\cellcolor[HTML]{FFFFC7}75.96\%} &
  \multicolumn{1}{c|}{\cellcolor[HTML]{FFCCC9}70.97\%} &
  \multicolumn{1}{c|}{\cellcolor[HTML]{9AFF99}80.37\%} &
  \multirow{-2}{*}{\textbf{Regs}} &
  \multicolumn{1}{c|}{-} &
  \multicolumn{1}{c|}{-} &
  \multicolumn{1}{c|}{-} &
  - \\ \cline{2-2} \cline{4-13} 
 &
  \multicolumn{1}{c|}{} &
  \multicolumn{1}{c|}{} &
  $10^{-6}$ &
  \multicolumn{1}{c|}{\cellcolor[HTML]{FFCE93}65.79\%} &
  \multicolumn{1}{c|}{\cellcolor[HTML]{FFFFC7}70.83\%} &
  \multicolumn{1}{c|}{\cellcolor[HTML]{FFCCC9}61.11\%} &
  \multicolumn{1}{c|}{\cellcolor[HTML]{9AFF99}74.86\%} &
   &
  \multicolumn{1}{c|}{\cellcolor[HTML]{FFFFC7}80.28\%} &
  \multicolumn{1}{c|}{\cellcolor[HTML]{FFCE93}77.10\%} &
  \multicolumn{1}{c|}{\cellcolor[HTML]{FFCCC9}65.94\%} &
  \cellcolor[HTML]{9AFF99}81.95\% \\ \cline{4-8} \cline{10-13} 
\multirow{-5}{*}{\textbf{Resnet18}} &
  \multicolumn{1}{c|}{\multirow{-2}{*}{\textbf{Weights}}} &
  \multicolumn{1}{c|}{\multirow{-4}{*}{\textbf{BER}}} &
  $10^{-5}$ &
  \multicolumn{1}{c|}{\cellcolor[HTML]{FFCE93}30.66\%} &
  \multicolumn{1}{c|}{\cellcolor[HTML]{FFFFC7}41.26\%} &
  \multicolumn{1}{c|}{\cellcolor[HTML]{FFCCC9}30.64\%} &
  \multicolumn{1}{c|}{\cellcolor[HTML]{9AFF99}43.92\%} &
  \multirow{-2}{*}{\textbf{FUs}} &
  \multicolumn{1}{c|}{\cellcolor[HTML]{FFFFC7}67.27\%} &
  \multicolumn{1}{c|}{\cellcolor[HTML]{FFCE93}50.35\%} &
  \multicolumn{1}{c|}{\cellcolor[HTML]{9AFF99}71.70\%} &
  \cellcolor[HTML]{FFCCC9}42.53\% \\ \hline
 &
  \multicolumn{3}{c|}{\textbf{Golden accuracy}} &
  \multicolumn{1}{c|}{98.00\%} &
  \multicolumn{1}{c|}{98.00\%} &
  \multicolumn{1}{c|}{98.25\%} &
  \multicolumn{1}{c|}{98.00\%} &
   &
  \multicolumn{1}{c|}{98.00\%} &
  \multicolumn{1}{c|}{98.00\%} &
  \multicolumn{1}{c|}{98.25\%} &
  98.00\% \\ \cline{2-13} 
 &
  \multicolumn{1}{c|}{} &
  \multicolumn{1}{c|}{} &
  $10^{-6}$ &
  \multicolumn{1}{c|}{\cellcolor[HTML]{FFCCC9}83.60\%} &
  \multicolumn{1}{c|}{\cellcolor[HTML]{9AFF99}95.65\%} &
  \multicolumn{1}{c|}{\cellcolor[HTML]{FFCE93}93.19\%} &
  \multicolumn{1}{c|}{\cellcolor[HTML]{FFFFC7}94.80\%} &
   &
  \multicolumn{1}{c|}{\cellcolor[HTML]{FFFFC7}98.00\%} &
  \multicolumn{1}{c|}{\cellcolor[HTML]{FFCCC9}97.10\%} &
  \multicolumn{1}{c|}{\cellcolor[HTML]{FFCE93}97.23\%} &
  \cellcolor[HTML]{FFFFC7}98.00\% \\ \cline{4-8} \cline{10-13} 
 &
  \multicolumn{1}{c|}{\multirow{-2}{*}{\textbf{Neurons}}} &
  \multicolumn{1}{c|}{} &
  $10^{-5}$ &
  \multicolumn{1}{c|}{\cellcolor[HTML]{FFCCC9}75.12\%} &
  \multicolumn{1}{c|}{\cellcolor[HTML]{9AFF99}93.96\%} &
  \multicolumn{1}{c|}{\cellcolor[HTML]{FFCE93}88.10\%} &
  \multicolumn{1}{c|}{\cellcolor[HTML]{FFFFC7}93.55\%} &
  \multirow{-2}{*}{\textbf{Regs}} &
  \multicolumn{1}{c|}{-} &
  \multicolumn{1}{c|}{-} &
  \multicolumn{1}{c|}{-} &
  - \\ \cline{2-2} \cline{4-13} 
 &
  \multicolumn{1}{c|}{} &
  \multicolumn{1}{c|}{} &
  $10^{-6}$ &
  \multicolumn{1}{c|}{\cellcolor[HTML]{FFCE93}95.37\%} &
  \multicolumn{1}{c|}{\cellcolor[HTML]{FFFFC7}97.30\%} &
  \multicolumn{1}{c|}{\cellcolor[HTML]{FFCCC9}94.70\%} &
  \multicolumn{1}{c|}{\cellcolor[HTML]{9AFF99}97.40\%} &
   &
  \multicolumn{1}{c|}{\cellcolor[HTML]{FFFFC7}93.55\%} &
  \multicolumn{1}{c|}{\cellcolor[HTML]{FFCE93}92.81\%} &
  \multicolumn{1}{c|}{\cellcolor[HTML]{FFCCC9}87.09\%} &
  \cellcolor[HTML]{9AFF99}95.96\% \\ \cline{4-8} \cline{10-13} 
\multirow{-5}{*}{\textbf{Lenet5}} &
  \multicolumn{1}{c|}{\multirow{-2}{*}{\textbf{Weights}}} &
  \multicolumn{1}{c|}{\multirow{-4}{*}{\textbf{BER}}} &
  $10^{-5}$ &
  \multicolumn{1}{c|}{\cellcolor[HTML]{FFCCC9}75.54\%} &
  \multicolumn{1}{c|}{\cellcolor[HTML]{9AFF99}95.01\%} &
  \multicolumn{1}{c|}{\cellcolor[HTML]{FFCE93}76.35\%} &
  \multicolumn{1}{c|}{\cellcolor[HTML]{FFFFC7}93.50\%} &
  \multirow{-2}{*}{\textbf{FUs}} &
  \multicolumn{1}{c|}{\cellcolor[HTML]{FFCCC9}86.20\%} &
  \multicolumn{1}{c|}{\cellcolor[HTML]{9AFF99}91.63\%} &
  \multicolumn{1}{c|}{\cellcolor[HTML]{FFCE93}90.51\%} &
  \cellcolor[HTML]{FFFFC7}91.42\% \\ \hline
  \label{tab:transposed_colored_complete_table}
\end{tabular}
\vspace{-7.5 mm}
\end{table*}
    In order to make a fair comparison of the results obtained from single stuck-at faults injected at ISA and APP, the $BER$ has been computed after performing ISA FIs according to the number of times that the injected fault is activated. The evaluations at ISA proved that the generated $BER$ was within the range of $[0$,$10^{-5}]$. To ensure comparability, APP FI campaigns were conducted with various BERs within the specified range. This approach allowed us to evaluate the accuracy of the applications across different abstraction levels. 
    

    Table \ref{tab:transposed_colored_complete_table} reports the DNN accuracy \textit{i)} when Neurons and Weights are corrupted at the APP by multiple bit-flips (that represent the corruption effect from individual stuck-at faults) with a BER assuming values within the 2 selected bins ($[0,$$10^{-6}$), [$10^{-6}$, $10^{-5}$]) and \textit{ii)} when single stuck-at faults affect FUs and Regs generating a BER value, which represents the activation of the error (BER, DNN level) on the columns. 

     In our evaluations, we observed that single stuck-at faults in RFs never produced BER errors in the range [$10^{-6}$, $10^{-5}$]. The error induced at the SW level with a high rate generates many errors that make the GPU hang or crush (i.e., DUEs).
     

    From a DNN's architecture standpoint, Lenet5 demonstrates higher accuracy resiliency to both ISA and APP evaluation abstractions with 
    a maximum degradation of 22.88\% w.r.t. the golden accuracy, thanks to its lower complexity.
    On the contrary, more elaborated DNNs, such as Mobilenet V2 and Resnet18, report $2\times$ of accuracy degradation (46.96\% and 42.16\%, respectively), w.r.t. the golden accuracy.

    More in detail, the most aggressive APP evaluation (i.e., when BER =$10^{-5}$) of Resnet18 and Lenet5 show reduced accuracy (never going above 30.64\% and 75.12\% w.r.t. the performance reported when single stuck-at faults at ISA have generated the same BER). Specifically, in the latter case, the reported accuracy never reaches more than 42.53\% and 86.20\%. On the other hand, the hardening strategies can raise those accuracy values to 80.37\% for Resnet18 and to 95.01\% when faults are injected at the DNN level. 
    On the contrary, the reported accuracy of Mobilenet V2 varies within a larger range when BER is set to $10^{-5}$ (i.e., [51.28\%, 83.63\%]) than ISA FIs, where the accuracy never goes above 77.9\%.

    Overall, we can observe that Adaptive Clipper achieves good performance only when it is implemented on Lenet5, reaching 93.96\% when neurons' outputs are corrupted and 95.01\% thanks to its light model size \ref{sec:only_single}. 
    Similarly, the implemented SW-hardening strategy outperforms the others when faults are injected in FUs with a 91.63\% of accuracy. 
    On the other hand, when the error induced by register corruption is around $10^{-6}$, the hardening of most SW-hardening strategies is neglected except for Ranger, which maintains the DNN accuracy to the baseline (98\%).
    Overall, Ranger performs well in most evaluations, especially when faults in neurons and weights impact Resnet18 by degrading the baseline model accuracy in only 6.42\% and 7.26\%, respectively. Nonetheless, during Resnet18 inference, under corrupted registers, none of the hardening strategies improves baseline DNN accuracy of 87.08\%. While Swap ReLU6 shows an accuracy of 71.70\% when FUs are affected by permanent faults. 
    
    Interestingly, in Mobilenet V2, Ranger is the most effective SW-hardening technique among all others in terms of accuracy degradation, when considering neurons (with 13.2\%), Regs (with 3.32\%), and FUs (with 12.81\%) corruption. This means that the fine-grained check at the convolutional block level performed by Ranger helps to fix the error induced by the corruption of the mentioned components that generate entries in the intermediate FMs that are outside of the valid range. 

        
    In conclusion, different abstraction levels yield different outcomes in FI campaigns. For example, Swap ReLU6 is ineffective under ISA FI but outperforms others in APP FI on Mobilenet V2. This proves that different FI strategies may provide significantly different results about the impact of hardening techniques on the DNN accuracy.  As a typical example, ISA FI with  $10^{-5}$ BER shows that hardened models worsen the accuracy w.r.t. the baseline, with up to a 24.74\% degradation in Resnet18.
    
\section{Conclusions}
    In this work, we evaluated and analyzed the effects of permanent faults at 2 abstraction levels (APP and ISA). We considered 3 SW  hardening techniques (Ranger, Adaptive Clipper, and Swap ReLU6).
    
    The results show that APP-level FI overestimates the beneficial impact of SW hardening techniques in terms of both reliability improvement and accuracy degradation. APP FIs tend to understimate the percentage of Critical SDCs and neglect that of DUEs. More in general, moving to a hardware-aware FI technique (hence, intrinsically more accurate) changes completely the ranking of the different hardening techniques: as an example, APP FI identifies Adaptive Clipper as the best solution in terms of reliability enhancement for Mobilenet V2, while for ISA FI the best solution is Swap ReLU6. 
    
    In future works, we plan to extend our analysis by exploring further hardening techniques, different underlying HW architectures, and more DNN models.  

\bibliographystyle{IEEEtran}
\bibliography{ref}

\end{document}